\documentclass[a4paper,nobib,notoc]{tufte-handout}
\usepackage[light,math]{kurier}
\usepackage[T1]{fontenc}
\usepackage{mathptmx}
\usepackage{soul}\setuldepth{article}

\usepackage{times}
\normalfont
\usepackage[T1]{fontenc}
\def\hb{\hbox to 11.5 cm{}}

\usepackage[utf8]{inputenc}
\usepackage[small]{caption}
\usepackage{graphicx}
\usepackage{amsmath}
\usepackage{amsthm}
\usepackage{amssymb}
\usepackage{booktabs}
\usepackage{relsize}
\usepackage{latexsym} 
\usepackage{algorithm}
\usepackage[noend]{algpseudocode}
\usepackage{xspace}
\usepackage{bm}

\usepackage{tikz}
\usetikzlibrary{arrows,automata,positioning,shapes,snakes}
\usetikzlibrary{calc,trees,positioning,arrows,chains,shapes.geometric,%
decorations.pathreplacing,decorations.pathmorphing,shapes,%
matrix,shapes.symbols,plotmarks,decorations.markings,shadows}
\usepackage{caption}
\usepackage{listings}

\usepackage[commentmarkup=uwave]{changes}
\definechangesauthor[color=magenta]{MT}
\definechangesauthor[color=blue]{FC}

\newcommand{\toprop}[2]{\ensuremath{\Phi}(#1, #2)}

\newcommand{\NNF}{{\tt NNF}}

\newcommand{\dDNNF}{{\tt d-DNNF}}

\newcommand{\gensem}{\sigma}
\newcommand{\setgenext}[2]{\mathcal{E}_{#1}(#2)}
\newcommand{\CF}{\ensuremath{\mathsf{CF}}}
\newcommand{\AD}{\ensuremath{\mathsf{AD}}}
\newcommand{\CO}{\ensuremath{\mathsf{CO}}}
\newcommand{\GR}{\ensuremath{\mathsf{GR}}}
\newcommand{\PR}{\ensuremath{\mathsf{PR}}}
\newcommand{\ST}{\ensuremath{\mathsf{ST}}}

\newcommand{\attackers}[1]{#1^{-}}
\newcommand{\attacked}[1]{#1^{+}}             

\newcommand{\tuple}[1]{\ensuremath{\langle #1 \rangle}}
\newcommand{\set}[1]{\ensuremath{\{#1\}}}

\newcommand{\argument}[1]{\ensuremath{\textrm{\textbf{#1}}}}
\newcommand{\arga}{\argument{a}}
\newcommand{\argb}{\argument{b}}
\newcommand{\argc}{\argument{c}}
\newcommand{\argd}{\argument{d}}

\newcommand{\AFname}{\ensuremath{AF}}
\newcommand{\setargs}{\ensuremath{\mathcal{A}}}
\newcommand{\setattacks}{\ensuremath{\mathcal{R}}}
\newcommand{\AF}[1]{\tuple{\setargs_{#1}, \setattacks_{#1}}}
\newcommand{\anAF}{\AF{}}
\newcommand{\anAFsymbol}{\ensuremath{\Gamma}}
\newcommand{\attacks}[2]{\ensuremath{#1 \rightarrow #2}}

\newcommand{\dungconffree}{conflict--free}
\newcommand{\dungacceptable}{acceptable}
\newcommand{\dungcharacteristic}{characteristic}
\newcommand{\dungadmissible}{admissible}
\newcommand{\dungcomplete}{complete}
\newcommand{\dunggrounded}{grounded}
\newcommand{\dungpreferred}{preferred}
\newcommand{\dungstable}{stable}

\newcommand{\charfun}[1]{\ensuremath{\mathcal{F}_{#1}}}

\newcommand{\aset}{\ensuremath{S}}

\newcommand{\labelfun}{\ensuremath{\rho}}
\newcommand{\weight}[1]{\ensuremath{w_{#1}}}

\newcommand{\ie}{i.e.,}
\newcommand{\eg}{e.g.,}

\newcommand{\mi}{\ensuremath{\supset}}

\newtheorem{theorem}{Theorem}

\newtheorem{proposition}[theorem]{Proposition}

\theoremstyle{definition}
\newtheorem{definition}[theorem]{Definition}
\newtheorem{example}[theorem]{Example}

\usepackage{relsize}
\usepackage{amsthm}
\usepackage{bm}

\newcommand{\dd}{\ensuremath{\mathrm{d}}}

\newcommand{\ev}{\ensuremath{\mathbb{E}}}

\newcommand{\Data}{\ensuremath{\mathcal{D}}}

\newcommand{\dbeta}{\ensuremath{\mathrm{Beta}}}

\newcommand{\noderv}[1]{\ensuremath{X_{#1}}}

\newcommand{\prior}[1]{\ensuremath{#1^{0}}}
\usepackage[]{authblk}

\newcommand{\fnms}[1]{#1}
\newcommand{\snm}[1]{#1}

\renewcommand{\subsubsection}[1]{\newthought{#1}}

\begin{document}

\title{Research Note on Uncertain Probabilities and Abstract Argumentation}

\author[$\dagger$]{\fnms{Pietro} \snm{Baroni}}
\author[$\dagger$,$\ddagger$]{\fnms{Federico} \snm{Cerutti}
\thanks{Corresponding Author; E-mail:
federico.cerutti@unibs.it.}}
\author[$\dagger$]{\fnms{Massimiliano} \snm{Giacomin}}
\author[$\star$]{\fnms{Lance M.} \snm{Kaplan}}
\author[$\#$]{\fnms{Murat} \snm{\c{S}ensoy}\thanks{The work was done prior to joining Amazon.}}

\affil[$\dagger$]{University of Brescia, Italy}
\affil[$\ddagger$]{Cardiff University, UK}
\affil[$\star$]{DEVCOM Army Research Laboratory, USA}
\affil[$\#$]{Amazon, Alexa AI, London, UK}

\date{}

\maketitle

\begin{abstract}
The sixth assessment of the international panel on climate change (IPCC) states that ``cumulative net CO2 emissions over the last decade (2010-2019) are about the same size as the
11 remaining carbon budget \textit{likely} to limit warming to 1.5C (\textit{medium confidence}).''
Such reports directly feed the public discourse, but nuances---such as the degree of
belief and of confidence---are often lost.
In this paper, we propose a formal account for allowing such degrees of belief and the associated confidence to be used to label arguments in abstract argumentation settings. Differently from other proposals in probabilistic argumentation, we  focus on the task of probabilistic inference over a chosen query building upon Sato's distribution semantics which has been already shown to encompass a variety of cases including the semantics of Bayesian networks. Borrowing from the vast literature on such semantics, we examine how such tasks can be dealt with in practice when considering uncertain probabilities, and discuss the connections with existing proposals for probabilistic argumentation.
\end{abstract}

\section{Introduction}
\label{sec:introduction}

The sixth assessment of the international panel on climate change (IPCC) states that ``cumulative net CO2 emissions over the last decade (2010-2019) are about the same size as the
11 remaining carbon budget \textit{likely} to limit warming to 1.5C (\textit{medium confidence})'' \cite[p. TS-16]{ipcc_ClimateChange2022_NaN}.
Such reports directly feed the public discourse, but nuances---such as the degree of
belief and of confidence---are often lost.

The degree of belief and the confidence in such a degree, however, refer to two different notions of uncertainty: \emph{aleatory} (or \emph{aleatoric}), and \emph{epistemic}  uncertainty \cite{HORA1996217}. Aleatory uncertainty refers to the variability in the outcome of an experiment which is due to inherently random effects (\eg\ flipping a fair coin): no additional source of information but Laplace's daemon\footnote{``An
intelligence that, at a given instant, could comprehend all the forces by
which nature is animated and the respective situation of the beings that
make it up'' \cite[p.2]{Laplace-prob}.} can reduce such a variability. Epistemic uncertainty refers to the epistemic state of the agent using the model, hence its lack of knowledge that---in principle---can be reduced on the basis of additional data samples (\eg\ being in the position to assess whether a coin is fair or not requires trials).

The ultimate goal of this research is to support human decision makers
in their tasks of reasoning in presence of aleatory and epistemic
uncertainty. 
Beta distributions (Section \ref{sec:beta}) can provide a compact representation of both
aleatory and epistemic uncertainty when dealing with binary
variables \cite{cerutti_EvidentialReasoningLearning_22}, \ie\ that can be only true or false, or---in
argumentation terms---arguments that can be accepted or not
  according to some acceptance criterion: in particular we focus on credulous acceptance. Moreover, beta distributions can be also translated
into expressions such as \textit{likely} with \textit{some confidence} \cite[p. 49]{josang_Subjectivelogic_16}.

For what concerns the reasoning under aleatory and epistemic uncertainty, most of the community's efforts focused on Bayesian networks and probabilistic logic programming under distribution semantics \cite{sato:iclp95,Fierens2015}, which we will illustrate in the form of algebraic model
counting \cite{kimmig_Algebraicmodelcounting_17} (Section \ref{sec:backreasoning}). In particular, \cite{Kaplan2018} looks at aleatory and epistemic uncertainty propagation in singly-connected Bayesian networks; \cite{cerutti_ProbabilisticLogicProgramming_19} provides efficient mechanisms for embedding aleatory and epistemic uncertainty in probabilistic logic programs; and finally \cite{cerutti_HandlingEpistemicAleatory_22} generalises the approach considering probabilistic circuits.

We then introduce the distribution semantics in abstract argumentation (Section \ref{sec:distribution}) at first considering only probabilities, with an in-depth discussion of similarities and differences with the constellation approach to probabilistic argumentation \cite{li_Relaxingindependenceassumptions_11,hunter_probabilisticapproachmodelling_13,hunter_Probabilisticqualificationattack_14}. We then show how the same semantics can handle beta distributions by leveraging recent work on probabilistic circuits \cite{cerutti_HandlingEpistemicAleatory_22}.
Section \ref{sec:conclusions} concludes the paper pencilling in a series of related research questions  that should be addressed
by this line of research.

\section{A Formal Account of Aleatory and Epistemic Uncertainty using Beta Distributions}
\label{sec:betareasoningback}

Scientific assessments of complex phenomena \cite{mastrandrea_IPCCAR5guidance_11} require to consider not only their likelihood but also the overall confidence in the exercise.
Let us suppose we know there is a coin that can be flipped, and a source of information is telling us that it is unlikely with low confidence that it will land on the head. This is a way to express uncertain probabilities, for instance because of limited observations: just observing a few times a coin flipping we can only have a low confidence on whether the coin is fair or not. 

\subsection{Representing Epistemic and Aleatory Uncertainty as Beta Distributions}
\label{sec:beta}

Epistemic and aleatory uncertainty can be jointly captured by a beta distribution, namely a distribution of possible probabilities. 
When facing a phenomenon with just two outcomes 
a \emph{complete dataset} $\Data$ is a sequence (allowing for repetitions) of examples, each being a vector of instantiations 
of 
independent Bernoulli distributions
with true but unknown parameter ${{ \pi}}$.
From this, the \emph{likelihood} is thus:
%
    $p(\Data \mid { \pi}) = \prod_{n=1}^{|\Data|} p({x}_n \mid { \pi}) = \prod_{n=1}^{N} { \pi}^{x_n} (1 - { \pi})^{1 - x_n}$
%
where ${x}_i$ represents the $i$-th example in the dataset \Data, that is assumed to hold either the value 1 or 0.

To develop a Bayesian analysis of the phenomenon, we can choose as prior the beta distribution, with parameters $\bm{\alpha} = \tuple{\alpha_x, \alpha_{\overline{x}}}$, $\alpha_x > 0$ and $\alpha_{\overline{x}} > 0$, that is conjugate to the Bernoulli:
%
    $\dbeta({ \pi} \mid \bm{\alpha}) = \frac{\Gamma(\alpha_x + \alpha_{\overline{x}})}{\Gamma(\alpha_x) \Gamma(\alpha_{\overline{x}})} { \pi}^{\alpha_x-1} (1 - { \pi})^{\alpha_{\overline{x}} -1}$
where
    $\Gamma(t) \equiv \int_0^{\infty} u^{t-1} e^{-u} \dd u$
is the \textit{gamma} function.

Considering a beta distributed prior $\dbeta(\pi \mid \prior{\bm{\alpha}})$ and the Bernoulli likelihood function, and given $|\Data|$ observations $\bm{m} = \tuple{m_x, m_{\overline{x}}}$ of $x$, viz., $m_x$ observations of $x=1$, $m_{\overline{x}}$ observations of $x = 0$, and $m_x + m_{\overline{x}} = |\Data|$:
    $\displaystyle{p({\pi} \mid \Data, \prior{\bm{\alpha}}) = \dbeta(\pi \mid \prior{\bm{\alpha}} + \bm{m})}$.

Thus, the parameters of a beta distribution can be considered pseudocounts \cite{murphy_Machinelearningprobabilistic_12} of \textit{pieces of evidence} for the two outcomes of a phenomenon, and the beta distribution itself can be seen as a representation of the uncertain probability associated with the phenomenon. Among the various priors, using $\prior{\bm{\alpha}} = \bm{1} = \tuple{1, 1}$ is equivalent to using the uniform distribution, which represents a non-informative prior that maximises entropy.

Given a beta-distributed random variable $X$, 
$s_X = \alpha_x + \alpha_{\bar{x}}$
 is its \emph{Dirichlet strength} and 
    $\mu_X = \frac{\alpha_x}{s_X}$
is its mean, which relates to the \emph{aleatory uncertainty}. 
The variance of a beta-distributed random variable---which associates
to the \emph{epistemic uncertainty}---$X$ is
$\sigma_X^2 = \frac{\mu_X (1 - \mu_X)}{s_X+1}.$

Following \cite[p. 49]{josang_Subjectivelogic_16} we can
represent fuzzy labels\footnote{The list of fuzzy labels for aleatory uncertainty is \cite[p. 49]{josang_Subjectivelogic_16}: \textit{absolutely not likely}, \textit{very unlikely}, \textit{unlikely}, \textit{somewhat unlikely}, \textit{chances about even}, \textit{somewhat likely}, \textit{likely}, \textit{very likely}, \textit{absolutely likely}. The list of fuzzy labels for epistemic uncertainty is \cite[p. 49]{josang_Subjectivelogic_16}: \textit{no confidence}, \textit{low confidence}, \textit{some confidence}, \textit{high confidence}, \textit{total confidence}.} such as \emph{likely} with \emph{low confidence} by means of beta distributions, by associating the first label to ranges of expected values (aleatory uncertainty) and the latter to ranges of variance values (epistemic uncertainty): the lower the confidence, the higher the variance. For instance, a proposition that is \emph{likely} true with \emph{some confidence} could be associated to a beta distribution $\dbeta(5.00, 1.50)$ as its expected value ($0.7692$) sits at the centre of the chosen range of values for the fuzzy label \emph{likely}, and analogously the variance ($0.0237$) for the fuzzy label \emph{some confidence}.

\subsection{Logical Reasoning with Beta Distributions}
\label{sec:backreasoning}

In \cite{cerutti_HandlingEpistemicAleatory_22} the authors provide a formal account for reasoning over probabilistic circuits which can be derived from propositional theories through knowledge compilation. In this setting, probabilistic inference is a special case of 
algebraic model counting (AMC) \cite{kimmig_Algebraicmodelcounting_17}, which generalises weighted
model counting (WMC) to the semiring setting and supports various
types of labels,  including numerical ones as used in WMC,  but also sets,
polynomials, Boolean formulae, and many more. The underlying mathematical structure is that of a
commutative semiring. 

A \emph{semiring} is a structure $\tuple{\mathcal{A}, \oplus, \otimes, 
e^{\oplus}, e^{\otimes}}$, where \emph{addition}~$\oplus$ and
\emph{multiplication}~$\otimes$ are associative binary operations over
the set~$\mathcal{A}$, $\oplus$~is commutative, $\otimes$~distributes
over~$\oplus$, $e^{\oplus}\in\mathcal{A}$ is the neutral element of~$\oplus$, $e^{\otimes}\in\mathcal{A}$ that of~$\otimes$,
and for all $a\in \mathcal{A}$, $e^{\oplus}\otimes a = a \otimes
e^{\oplus} = e^{\oplus}$. In a \emph{commutative semiring}, $\otimes$~is
commutative as well.

Algebraic model counting is now defined \cite{kimmig_Algebraicmodelcounting_17} as follows. Given:
  a \emph{propositional logic theory} $T$ over a set of
    variables $\mathcal{V}$, 
   a \emph{commutative semiring} $\tuple{\mathcal{A},\oplus,\otimes,   e^{\oplus},e^{\otimes}}$, and 
  a \emph{labelling function} $\labelfun : \mathcal{L} \rightarrow \mathcal{A}$, mapping literals $\mathcal{L}$ of the variables in $\mathcal{V}$ to elements of the semiring set $\mathcal{A}$, 
compute
\begin{equation} \label{eq:amc}
\operatorname{\mathbf{A}}(T)  = \bigoplus_{I\in \mathcal{M}(T)} \bigotimes_{l \in I} \labelfun(l),
\end{equation}
where $\mathcal{M}(T)$ denotes the set of models of~$T$.

\subsubsection{Probabilistic Inferences}
Among others, AMC generalises the task of
probabilistic inference according to \cite{sato:iclp95}'s semantics (\textbf{PROB}), \cite[Thm. 1]{kimmig_Algebraicmodelcounting_17}.
A \emph{query} $q$ is a finite set of algebraic literals  $q\subseteq \mathcal{L}$. We denote the set of interpretations where the query is true by $\mathcal{I}(q)$,
\begin{equation}
  \label{eq:interpretationquery}
\mathcal{I}(q) = \{I~|~I\in \mathcal{M}(T) ~\land~ q \subseteq I\}
\end{equation}

The 
label
of a query $q$ is defined\footnote{Albeit $\operatorname{\mathbf{A}}$ has been introduced to operate over propositional logic theories, with a small abuse of notation we use it also for a finite set of literals, \ie\ \emph{query}, and a set of interpretations.} as the label of $\mathcal{I}(q)$,
\begin{equation}
\operatorname{\mathbf{A}}(q)  =  \operatorname{\mathbf{A}}(\mathcal{I}(q)) = \bigoplus_{I\in \mathcal{I}(q)}\bigotimes_{l\in I}\rho(l).\label{eq:q_int}
\end{equation}
As both operators are commutative and associative, the label is independent of the order of both literals and interpretations.

In the case of probabilities as labels, \ie\ $\rho(\cdot) \in [0,1]$, \eqref{eq:semiringprobability} presents the AMC parametrisation for handling \textbf{PROB} of 
queries:
\begin{fullwidth}
\begin{equation}
    \label{eq:semiringprobability}
    \begin{array}{l p{.7cm} l p{.7cm} l}
         \mathcal{A} = \mathbb{R}_{\geq 0} & &
         a ~\oplus~ b = a + b & &
         a ~\otimes~ b = a \cdot b\\
         e^\oplus = 0 & &
         e^{\otimes} = 1 & & 
         \rho(f) \in [0,1] \mbox{ and }
         \overline{\rho(f)} = \rho(\lnot f) = 1 - \rho(f)\\
    \end{array}
\end{equation}
\end{fullwidth}

A na\"ive implementation of \eqref{eq:amc} 
is clearly exponential: \cite{Darwiche2004} introduced the first method for deriving tractable circuits (\dDNNF s) that allow polytime algorithms for clausal
entailment, model counting and enumeration. 
In particular, we can exploit the succinctness results of 
the knowledge compilation map by \cite{darwiche_KnowledgeCompilationMap_02}, where an overview of succinctness relationships between various
types of  circuits is provided. Instead of focusing on classical, flat target compilation languages based on conjunctive or disjunctive normal forms, \cite{darwiche_KnowledgeCompilationMap_02} 
considers a richer, nested class based on representing propositional sentences using directed acyclic
graphs: \NNF s.
A sentence in \emph{negation normal form} (\NNF) over a set of propositional variables $\mathcal{V}$
is a rooted, directed acyclic graph where each leaf node is labeled
with true ($\top$), false ($\bot$), or a literal of a variable in~$\mathcal{V}$, and
each internal node with disjunction ($\vee$) or conjunction ($\wedge$).

An \NNF\ is \emph{decomposable} if for each conjunction
node~$\bigwedge_{i=1}^n\phi_i$, no two children~$\phi_i$ and~$\phi_j$
share any variable. 
An \NNF\ is \emph{deterministic} if for each disjunction
node~$\bigvee_{i=1}^n\phi_i$, each pair of different 
children~$\phi_i$ and~$\phi_j$ is logically contradictory, that is
$\phi_i \land \phi_j \models \bot$ for $i \neq j$. In other terms,
only one child can be true at any time.
An \NNF\ is \emph{smooth} when $\mathit{vars}(\phi_i) =
\mathit{vars}(\phi_j)$ for any two children $\phi_i$ and $\phi_j$ of a
disjunction node, where $\mathit{vars}(x)$ denotes
all propositional variables that appear in the subgraph rooted at
$x$.

The function \textsc{Eval} specified in Algorithm~\ref{alg:eval} \emph{evaluates} an \NNF\ circuit for a commutative semiring $(\mathcal{A},\oplus,\otimes, e^{\oplus},e^{\otimes})$ and labelling function $\labelfun$.
Evaluating an \NNF\ representation~$N_T$ of a propositional theory~$T$
for a  semiring $(\mathcal{A},\oplus,\otimes,e^{\oplus},e^{\otimes})$
and labelling function~$\labelfun$   is a
\emph{sound AMC computation} iff
\textsc{Eval}$(N_T,\oplus,\otimes,e^{\oplus},e^{\otimes},\labelfun)
=\operatorname{\mathbf{A}}(T)$.

Following the intuition in \cite{darwiche_differentialapproach_03},
the same circuit can be reused for multiple queries. For instance,
from \eqref{eq:interpretationquery} and \eqref{eq:q_int}, the very
same circuit used for computing $\operatorname{\mathbf{A}}(T)$ can be used
for computing the probability of a query $q$ 
by enforcing that the leaf
associated to $\lnot q$ carries no weight in the final
computation. For further details, see \cite{darwiche_differentialapproach_03,Fierens2015}.

\begin{algorithm}[t]
  \caption[\textsc{Label}]{Evaluating an \NNF\ circuit $N$ for a
    commutative semiring $(\mathcal{A},\oplus,\otimes,
    e^{\oplus},e^{\otimes})$ and labelling function $\labelfun$.}
\label{alg:eval}
\begin{algorithmic}[1]
\Procedure{\textsc{Eval}}{$N,\oplus,\otimes,e^{\oplus},e^{\otimes},\labelfun$}
\State \textbf{if} $N$ is a true node $\top$ \textbf{then} \textbf{return} $e^{\otimes}$
\State \textbf{if} $N$ is a false node $\bot$ \textbf{then} \textbf{return} $e^{\oplus}$
\State \textbf{if} $N$ is a literal node $l$ \textbf{then} \textbf{return} $\labelfun(l)$
\If{$N$ is a disjunction $\bigvee_{i=1}^mN_i$} \State \textbf{return} $\bigoplus_{i=1}^m$ \textsc{Eval}($N_i,\oplus,\otimes,e^{\oplus},e^{\otimes},\labelfun$) \EndIf
\If{$N$ is a conjunction $\bigwedge_{i=1}^mN_i$} \State \textbf{return} $\bigotimes_{i=1}^m$ \textsc{Eval}($N_i,\oplus,\otimes,e^{\oplus},e^{\otimes},\labelfun$) \EndIf
\EndProcedure
\end{algorithmic}
\end{algorithm}

\subsubsection{Inferences with Beta Distributions}
\label{sec:mach22}
In order to use beta distributions as labels, in \cite{cerutti_HandlingEpistemicAleatory_22}, the authors provide operators that expand on \eqref{eq:semiringprobability} for what concerns expected values of the random variables, and approximating the resulting variance using Taylor approximations. They
require as input also the matrix of covariance between such random variables, which captures their second-order dependencies.  
In their
proposal, $e^\oplus$ is a beta distribution with expected value 0
and 0 variance. $e^\otimes$ is a beta distribution with expected value 1
and 0 variance. 
 
Looking at Algorithm \ref{alg:eval}, there is then the need to introduce specific operators for $\bigoplus$ and $\bigotimes$, which are used when facing respectively a disjunction or a conjunction node. In the case of a disjunction $n$ over $C$ nodes---its children---the operators proposed in \cite{cerutti_HandlingEpistemicAleatory_22} return a beta distribution whose expected value is the sum of the expected values of its children, hence
$\ev[\noderv{n}] = \sum_{c\in C} \ev[\noderv{c}]$.
In the case of a conjunction, the operator returns a beta distribution whose expected value is approximated to the product of the expected values of its children, ${\ev[\noderv{n}]} \simeq {\Pi(\ev[\bm{\noderv{C}}])}$, as it results from a Taylor approximation. The operators proposed in \cite{cerutti_HandlingEpistemicAleatory_22} also try to faithfully approximate the variance of the resulting beta distributions, for both disjunction and conjunction nodes, by manipulating the covariances between each of their children and the other nodes in the circuit.

\section{Distribution Semantics in Abstract Argumentation}
\label{sec:distribution}

In Section \ref{sec:betareasoningback} we briefly recalled techniques  for  reasoning about epistemic and aleatory uncertainty represented through beta distributions. We now turn our attention to the issue of supporting argumentative-probabilistic reasoning.

\subsection{Background in Abstract Argumentation}


An \emph{argumentation framework} (\AFname) \cite{dung_acceptabilityarguments_95} is a pair $\anAFsymbol = \anAF$ 
where $\setargs$ is a set of arguments and $\setattacks \subseteq \setargs \times \setargs$.
We say that \argb{} \emph{attacks} \arga{} iff $\tuple{\argb,\arga} \in \setattacks$, also denoted as $\attacks{\argb}{\arga}$.
The set of attackers of an argument $\arga$ is denoted as 
$\attackers{\arga} \triangleq \set{\argb : \attacks{\argb}{\arga}}$,
the set of arguments attacked by $\arga$ is denoted as
$\attacked{\arga} \triangleq \set{\argb : \attacks{\arga}{\argb}}$.
Analogously, we can define the set of arguments attacked by a set of arguments $E \subseteq \setargs$ as 
$\attacked{E} \triangleq \set{\argb \mid \exists \arga \in E, \attacks{\arga}{\argb}}$.


The basic properties of \dungconffree ness, acceptability, and admissibility of a set of arguments 
are fundamental for the definition of argumentation semantics.
%
Given an $\AFname$ $\anAFsymbol = \anAF$:
\begin{itemize}
  \item a set $\aset \subseteq \setargs$ is a \emph{\dungconffree} set of $\anAFsymbol$
        if $\nexists~ \arga, \argb \in \aset$ s.t. $\attacks{\arga}{\argb}$;
  \item an argument $\arga \in \setargs$ is \emph{\dungacceptable} in $\anAFsymbol$ with respect to a set $\aset \subseteq \setargs$  
        if $\forall \argb \in \setargs$ s.t. $\attacks{\argb}{\arga}$, 
        $\exists~ \argc \in \aset$ s.t. $\attacks{\argc}{\argb}$;
  \item the function $\charfun{\anAFsymbol}: 2^{\setargs} \rightarrow 2^{\setargs}$ such that
        $\charfun{\anAFsymbol}(\aset) = \set{\arga \mid \arga \mbox{ is } \mbox{\dungacceptable} \mbox{ in $\anAFsymbol$} \mbox{ w.r.t. } \aset}$
        is called the \emph{\dungcharacteristic{} function} of \anAFsymbol.

\end{itemize}

An argumentation semantics $\gensem$ prescribes for any \AFname{}
$\anAFsymbol$ a set of \emph{extensions}, denoted as
$\setgenext{\gensem}{\anAFsymbol}$, i.e., a set of sets of arguments
satisfying the conditions dictated by $\gensem$. 

Given an $\AFname$ $\anAFsymbol = \anAF$, a set $\aset \subseteq \setargs$ is:
\begin{itemize}
    \item an \emph{admissible
        extension} of $\anAFsymbol$, \ie{} $\aset \in
      \setgenext{\AD}{\anAFsymbol}$, iff $\aset$ is an admissible
      set of $\anAFsymbol$, \ie{} $\aset \subseteq \charfun{\anAFsymbol}(\aset)$;
    \item a \emph{complete
        extension} of $\anAFsymbol$, \ie{} $\aset \in
      \setgenext{\CO}{\anAFsymbol}$, iff 
      $\aset \in
      \setgenext{\CF}{\anAFsymbol}$ and 
      $\aset =
      \charfun{\anAFsymbol}(\aset)$;
    \item the \emph{\dunggrounded{} extension} of $\anAFsymbol$, 
             \ie{} $\aset \in \setgenext{\GR}{\anAFsymbol}$, 
              iff $\aset$ is the minimal (w.r.t. set inclusion) \dungcomplete{} extension of  $\anAFsymbol$;
    \item a \emph{\dungstable{} extension} of $\anAFsymbol$, \ie{} $\aset \in \setgenext{\ST}{\anAFsymbol}$, iff $\aset$ is a conflict-free set of $\anAFsymbol$ and  $\aset\cup\attacked{\aset}=\setargs$;
    \item a \emph{\dungpreferred{} extension} of $\anAFsymbol$, 
             \ie{} $\aset \in \setgenext{\PR}{\anAFsymbol}$, 
              iff $\aset$ is a maximal (w.r.t. set inclusion) \dungadmissible{} set of  $\anAFsymbol$.
\end{itemize}

Given an argumentation framework $\anAFsymbol$, an argument $\arga$, and
a semantics $\gensem$,
$\arga$ is said to be \emph{credulously accepted} according to $\gensem$
if $\exists \aset \in \setgenext{\gensem}{\anAFsymbol}$ such that $\arga \in \aset$.

In the following we make use of the fact that given an argumentation
framework $\anAFsymbol$ and a semantics $\gensem$, it is possible to
derive a propositional theory whose models are
$\setgenext{\gensem}{\anAFsymbol}$.
For instance, from \cite{besnard_Checkingacceptabilityset_04} an
argumentation framework can be transformed into a propositional theory
whose models are isomorphic to the admissible sets. In particular, from \cite[Prop. 6]{besnard_Checkingacceptabilityset_04}, such a formula can be:
\begin{equation}
\label{eq:admissibility}
    \bigwedge_{\arga \in \setargs} 
    \left( 
        \left( \arga \mi \bigwedge_{\argb \in \set{\arga}^-}\lnot \argb \right) \land 
        \left( \arga \mi \bigwedge_{\argb \in \set{\arga}^-} \bigvee_{\argc \in \set{\argb}^-} \argc \right)
    \right)
  \end{equation}

Given an argumentation framework $\anAFsymbol$ and a semantics
$\gensem$, let us denote with $\toprop{\anAFsymbol}{\gensem}$ an arbitrary
propositional theory whose models are isomorphic to $\setgenext{\anAFsymbol}{\gensem}$.
Please note that we do not claim that such a theory can always be
derived in polynomial time from the definition of the argumentation
framework.

To illustrate our proposal, let us consider the following running
example loosely connected 
to the IPCC sixth's assessment \cite[p. TS-16]{ipcc_ClimateChange2022_NaN}
(cf. Section \ref{sec:introduction}). 

 \begin{example}
   \label{ex:ex}
  Let consider the argument $\argd$ stating that 
  ``from the collected evidence, we conclude that
  cumulative net CO2 emissions over the last decade (2010-2019) are about the same size as the
11 remaining carbon budget to limit warming to 1.5C.''
  Let us assume that such a conclusion involved
  assessing three further arguments, thus giving rise to 
the $\anAFsymbol_E = \tuple{\setargs_E, \setattacks_E}$
  argumentation framework where $\setargs_E = \set{\arga, \argb, \argc, \argd}$ and $\setattacks_E = \set{\attacks{\arga}{\argc}, \attacks{\argb}{\argc}, \attacks{\argc}{\argd}}$. 
  Given the sensitivity and complexity of the topic, we
  refrain from instantiating such additional arguments with natural
  language text for avoiding unnecessary debate over what should be
  just an illustrative example.

  From \eqref{eq:admissibility}, $\toprop{\anAFsymbol_E}{\AD} = (\argc
  \mi (\lnot \arga)) \land (\argc \mi (\lnot \argb)) \land (\argd \mi
  (\arga \lor \argb)) \land (\lnot \argc)$.
\end{example}

Figure \ref{fig:sdd} shows a smooth, deterministic, and decomposable
\NNF\ equivalent to $\toprop{\anAFsymbol_E}{\AD}$ generated using the method proposed in \cite{Darwiche2011}.

\begin{figure}
  \resizebox{\textwidth}{!}{%
    \begin{tikzpicture}[>=latex,line join=bevel,]
\node (1) at (99.0bp,18.0bp) [draw,rectangle,dotted,align=center] {\begin{tabular}{c}{\footnotesize $1$} \\[5pt]$\overline{\textbf{a}}$\\  
\end{tabular}};
  \node (2) at (27.0bp,18.0bp) [draw,rectangle,dotted,align=center] {\begin{tabular}{c}{\footnotesize $2$} \\[5pt]$\textbf{b}$\\  
\end{tabular}};
  \node (3) at (52.0bp,90.0bp) [draw,ellipse,dotted] {\begin{tabular}{c}{\footnotesize 3 }\\                             {\Large $\land$ }\end{tabular}};
  \node (4) at (171.0bp,18.0bp) [draw,rectangle,dotted,align=center] {\begin{tabular}{c}{\footnotesize $4$} \\[5pt]$\overline{\textbf{b}}$\\  
\end{tabular}};
  \node (5) at (124.0bp,90.0bp) [draw,ellipse,dotted] {\begin{tabular}{c}{\footnotesize 5 }\\                             {\Large $\lor$ }\end{tabular}};
  \node (6) at (243.0bp,18.0bp) [draw,rectangle,dotted,align=center] {\begin{tabular}{c}{\footnotesize $6$} \\[5pt]$\textbf{a}$\\  
\end{tabular}};
  \node (7) at (315.0bp,18.0bp) [draw,rectangle,dotted,align=center] {\begin{tabular}{c}{\footnotesize $7$} \\[5pt]$\top$\\  
\end{tabular}};
  \node (8) at (268.0bp,90.0bp) [draw,ellipse,dotted] {\begin{tabular}{c}{\footnotesize 8 }\\                             {\Large $\land$ }\end{tabular}};
  \node (9) at (124.0bp,162.0bp) [draw,ellipse,dotted] {\begin{tabular}{c}{\footnotesize 9 }\\                             {\Large $\land$ }\end{tabular}};
  \node (10) at (124.0bp,234.0bp) [draw,ellipse,dotted] {\begin{tabular}{c}{\footnotesize 10 }\\                             {\Large $\lor$ }\end{tabular}};
  \node (11) at (459.0bp,18.0bp) [draw,rectangle,dotted,align=center] {\begin{tabular}{c}{\footnotesize $11$} \\[5pt]$\textbf{d}$\\  
\end{tabular}};
  \node (12) at (251.0bp,306.0bp) [draw,ellipse,dotted] {\begin{tabular}{c}{\footnotesize 12 }\\                             {\Large $\land$ }\end{tabular}};
  \node (13) at (196.0bp,90.0bp) [draw,ellipse,dotted] {\begin{tabular}{c}{\footnotesize 13 }\\                             {\Large $\lor$ }\end{tabular}};
  \node (14) at (387.0bp,18.0bp) [draw,rectangle,dotted,align=center] {\begin{tabular}{c}{\footnotesize $14$} \\[5pt]$\overline{\textbf{d}}$\\  
\end{tabular}};
  \node (15) at (340.0bp,90.0bp) [draw,ellipse,dotted] {\begin{tabular}{c}{\footnotesize 15 }\\                             {\Large $\land$ }\end{tabular}};
  \node (16) at (196.0bp,162.0bp) [draw,ellipse,dotted] {\begin{tabular}{c}{\footnotesize 16 }\\                             {\Large $\land$ }\end{tabular}};
  \node (17) at (223.0bp,378.0bp) [draw,ellipse,dotted] {\begin{tabular}{c}{\footnotesize 17 }\\                             {\Large $\lor$ }\end{tabular}};
  \node (18) at (295.0bp,378.0bp) [draw,rectangle,dotted,align=center] {\begin{tabular}{c}{\footnotesize $18$} \\[5pt]$\overline{\textbf{c}}$\\  
\end{tabular}};
  \node (19) at (295.0bp,450.0bp) [draw,ellipse,dotted] {\begin{tabular}{c}{\footnotesize 19 }\\                             {\Large $\land$ }\end{tabular}};
  \node (20) at (412.0bp,90.0bp) [draw,ellipse,dotted] {\begin{tabular}{c}{\footnotesize 20 }\\                             {\Large $\lor$ }\end{tabular}};
  \node (21) at (531.0bp,18.0bp) [draw,rectangle,dotted,align=center] {\begin{tabular}{c}{\footnotesize $21$} \\[5pt]$\textbf{c}$\\  
\end{tabular}};
  \node (22) at (603.0bp,18.0bp) [draw,rectangle,dotted,align=center] {\begin{tabular}{c}{\footnotesize $22$} \\[5pt]$\bot$\\  
\end{tabular}};
  \node (23) at (531.0bp,90.0bp) [draw,ellipse,dotted] {\begin{tabular}{c}{\footnotesize 23 }\\                             {\Large $\land$ }\end{tabular}};
  \node (24) at (350.0bp,162.0bp) [draw,ellipse,dotted] {\begin{tabular}{c}{\footnotesize 24 }\\                             {\Large $\land$ }\end{tabular}};
  \node (25) at (322.0bp,522.0bp) [draw,ellipse,dotted] {\begin{tabular}{c}{\footnotesize 25 }\\                             {\Large $\lor$ }\end{tabular}};
  \draw [->,solid] (3) ..controls (68.305bp,64.716bp) and (75.368bp,54.196bp)  .. (1);
  \draw [->,solid] (3) ..controls (43.264bp,64.539bp) and (39.832bp,54.929bp)  .. (2);
  \draw [->,solid] (5) ..controls (93.103bp,66.703bp) and (74.915bp,53.578bp)  .. (2);
  \draw [->,solid] (5) ..controls (140.31bp,64.716bp) and (147.37bp,54.196bp)  .. (4);
  \draw [->,solid] (8) ..controls (259.26bp,64.539bp) and (255.83bp,54.929bp)  .. (6);
  \draw [->,solid] (8) ..controls (284.31bp,64.716bp) and (291.37bp,54.196bp)  .. (7);
  \draw [->,solid] (9) ..controls (124.0bp,135.98bp) and (124.0bp,126.71bp)  .. (5);
  \draw [->,solid] (9) ..controls (169.6bp,138.83bp) and (209.72bp,119.33bp)  .. (8);
  \draw [->,solid] (10) ..controls (104.56bp,207.05bp) and (95.022bp,193.15bp)  .. (88.0bp,180.0bp) .. controls (77.185bp,159.75bp) and (67.587bp,135.53bp)  .. (3);
  \draw [->,solid] (10) ..controls (124.0bp,207.98bp) and (124.0bp,198.71bp)  .. (9);
  \draw [->,solid] (12) ..controls (210.01bp,282.41bp) and (177.16bp,264.3bp)  .. (10);
  \draw [->,solid] (12) ..controls (311.6bp,266.23bp) and (409.02bp,196.01bp)  .. (448.0bp,108.0bp) .. controls (456.6bp,88.576bp) and (459.16bp,64.573bp)  .. (11);
  \draw [->,solid] (13) ..controls (165.1bp,66.703bp) and (146.91bp,53.578bp)  .. (1);
  \draw [->,solid] (13) ..controls (212.31bp,64.716bp) and (219.37bp,54.196bp)  .. (6);
  \draw [->,solid] (15) ..controls (331.26bp,64.539bp) and (327.83bp,54.929bp)  .. (7);
  \draw [->,solid] (15) ..controls (356.31bp,64.716bp) and (363.37bp,54.196bp)  .. (14);
  \draw [->,solid] (16) ..controls (171.25bp,136.94bp) and (157.48bp,123.55bp)  .. (5);
  \draw [->,solid] (16) ..controls (196.0bp,135.98bp) and (196.0bp,126.71bp)  .. (13);
  \draw [->,solid] (16) ..controls (241.6bp,138.83bp) and (281.72bp,119.33bp)  .. (15);
  \draw [->,solid] (17) ..controls (232.91bp,352.22bp) and (236.94bp,342.14bp)  .. (12);
  \draw [->,solid] (17) ..controls (218.65bp,349.47bp) and (216.61bp,335.99bp)  .. (215.0bp,324.0bp) .. controls (208.7bp,277.14bp) and (202.49bp,222.44bp)  .. (16);
  \draw [->,solid] (19) ..controls (270.25bp,424.94bp) and (256.48bp,411.55bp)  .. (17);
  \draw [->,solid] (19) ..controls (295.0bp,423.98bp) and (295.0bp,414.71bp)  .. (18);
  \draw [->,solid] (20) ..controls (428.31bp,64.716bp) and (435.37bp,54.196bp)  .. (11);
  \draw [->,solid] (20) ..controls (403.26bp,64.539bp) and (399.83bp,54.929bp)  .. (14);
  \draw [->,solid] (23) ..controls (531.0bp,63.983bp) and (531.0bp,54.712bp)  .. (21);
  \draw [->,solid] (23) ..controls (554.92bp,65.746bp) and (567.3bp,53.71bp)  .. (22);
  \draw [->,solid] (24) ..controls (287.84bp,146.88bp) and (217.6bp,129.46bp)  .. (160.0bp,108.0bp) .. controls (158.15bp,107.31bp) and (156.27bp,106.57bp)  .. (5);
  \draw [->,solid] (24) ..controls (304.47bp,140.83bp) and (265.45bp,123.41bp)  .. (232.0bp,108.0bp) .. controls (230.21bp,107.18bp) and (228.37bp,106.32bp)  .. (13);
  \draw [->,solid] (24) ..controls (371.53bp,136.69bp) and (382.56bp,124.24bp)  .. (20);
  \draw [->,solid] (24) ..controls (404.75bp,139.83bp) and (462.47bp,117.5bp)  .. (23);
  \draw [->,solid] (25) ..controls (312.49bp,496.34bp) and (308.67bp,486.43bp)  .. (19);
  \draw [->,solid] (25) ..controls (335.69bp,477.79bp) and (350.0bp,424.98bp)  .. (350.0bp,379.0bp) .. controls (350.0bp,379.0bp) and (350.0bp,379.0bp)  .. (350.0bp,305.0bp) .. controls (350.0bp,265.0bp) and (350.0bp,218.65bp)  .. (24);
\end{tikzpicture}%
  }
  \caption{Smooth, decomposable, and deterministic \NNF\ of
    $\toprop{\anAFsymbol_E}{\AD}$}
  \label{fig:sdd}
\end{figure}

\subsection{Algebraic Model Counting and Abstract Argumentation}

Before addressing the case of uncertain probabilities represented through beta distributions, let us discuss probabilistic inferences over argumentation frameworks considering simple probabilities and thus the very same setting illustrated in \eqref{eq:semiringprobability}.

A \emph{probabilistic graph} is a tuple $\tuple{\setargs, \setattacks,
  \rho}$ where $\tuple{\setargs, \setattacks}$ is an abstract
argumentation framework and $\rho: \setargs \to [0,1]$.
Given a probabilistic graph $\tuple{\setargs, \setattacks,
  \rho}$ and a set of arguments $\aset \subseteq \setargs$, the
probability of such a set of arguments is:
\begin{equation}
  \label{eq:probset}
  \rho(\aset) = \bigotimes_{\arga \in \aset} \rho(\arga) ~\otimes~ \bigotimes_{\arga \in \setargs\setminus\aset} \overline{\rho(\arga)} 
\end{equation}

While so far we share definitions with the constellation approach to probabilistic argumentation
\cite{li_Relaxingindependenceassumptions_11,hunter_probabilisticapproachmodelling_13,hunter_Probabilisticqualificationattack_14}
the probabilistic
inference according to \cite{sato:iclp95}'s semantics of a queried
argument $\arga$ as the sum of the probabilities of $\gensem$'s extensions\footnote{While in this paper we focus on the semantics
  introduced in Dung's paper \cite{dung_acceptabilityarguments_95}, Definition \ref{def:probquery}
  can apply to any semantics introduced over abstract
  argumentation framework that returns set(s) of acceptable arguments.}  in
which $\arga$ is present.\footnote{Definition \ref{def:probquery} generalises the semantics provided in \cite{totis_SMProbLogStableModel_21a}, where the authors restrict to the \CF\ case only.}

\begin{definition}
  \label{def:probquery}
  Given a probabilistic graph $\tuple{\setargs, \setattacks,
    \rho}$ over an argumentation framework $\anAFsymbol =
  \tuple{\setargs, \setattacks}$,  an argument $\arga \in \setargs$, and a semantics $\gensem
\in \set{\CF, \AD, \CO, \GR, \ST, \PR}$,
\begin{equation}
  \label{eq:probarg}
  \mbox{\textbf{PROB}}(\arga, \gensem, \anAFsymbol) = \bigoplus_{\set{E \in \setgenext{\gensem}{\anAFsymbol} ~\mid~ \arga \in E}} \rho(E).
\end{equation}
\end{definition}

We can now demonstrate that $\mbox{\textbf{PROB}}(\arga, \gensem,
\anAFsymbol)$ is an instance of AMC.

\begin{theorem}
  \label{thm:amcarg}
  Given a probabilistic graph $\tuple{\setargs, \setattacks, \rho}$
  over an argumentation framework $\anAFsymbol = \tuple{\setargs,
    \setattacks}$, and given $\arga \in \setargs$, 
  AMC generalises $\mbox{\textbf{PROB}}(\arga, \gensem,
  \anAFsymbol)$.

  \begin{proof}
    \cite{besnard_Checkingacceptabilityset_04} already provided
    $\toprop{\anAFsymbol}{\gensem'}$ for $\gensem' \in \set{\CF, \AD, \CO,
      \ST}$.
    It is however always possible to derive a propositional theory $\toprop{\anAFsymbol}{\gensem}$
    whose set of models is isomorphic to
    the set of extensions. For instance:
      $\toprop{\anAFsymbol}{\gensem} = \bigvee_{E \in \setgenext{\gensem}{\anAFsymbol}} ~~\left( \bigwedge_{\arga \in E} \arga ~\land~ \bigwedge_{\arga \in \setargs\setminus E} \lnot \arga 
      \right).$
%
    It is therefore immediate to see that, using the parametrisation
     \eqref{eq:semiringprobability}, \eqref{eq:q_int} over a theory $T$ is equivalent to
     \eqref{eq:probarg} when $T = \toprop{\anAFsymbol}{\gensem}$.
  \end{proof}
\end{theorem}

As discussed in the proof of Theorem \ref{thm:amcarg}, we do not make claims on how to derive a propositional theory whose set of models is isomorphic to the set of extensions for a given semantics. Moreover, since grounded semantics (\GR) returns a single extension, seeing its distribution semantics through the lenses of AMC might be an unnecessary complication. For what concerns preferred semantics ($\PR$), in addition to the trivial proposal suggested in the proof of Theorem \ref{thm:amcarg}---which requires the presence of an algorithm for computing $\setgenext{\anAFsymbol}{\PR}$ from an argumentation framework $\anAFsymbol$---in future work we will explore further its connection with MaxSAT problems, already investigated, among others, in \cite{faber_SolvingSetOptimization_16} and the connection between AMC and weighted MaxSAT, see for instance \cite{pipatsrisawat_SolvingWeightedMaxSAT_08}.

To show how Definition \ref{def:probquery} diverges from the constellation approach to probabilistic argumentation \cite{li_Relaxingindependenceassumptions_11,hunter_probabilisticapproachmodelling_13,hunter_Probabilisticqualificationattack_14}, let us recall its main definitions.

Given 
$\anAFsymbol = \tuple{\setargs, \setattacks}$ 
and     
$\anAFsymbol' = \tuple{\setargs', \setattacks'}$,
$\anAFsymbol'$ is a subgraph of $\anAFsymbol$, \ie\ $\anAFsymbol' \sqsubseteq \anAFsymbol$, iff $\setargs' \subseteq \setargs$ and $\setattacks' = \set{(\arga, \argb) \in \setattacks \mid \arga, \argb \in \setargs'}$. Let $\wp(\anAFsymbol) = \set{\anAFsymbol' \mid \anAFsymbol' \sqsubseteq \anAFsymbol}$.

Under independence assumptions, we can compute the probability of subgraphs as follows. Let $\anAFsymbol'  = \tuple{\setargs', \setattacks'} \sqsubseteq \anAFsymbol = \tuple{\setargs, \setattacks}$,
\begin{equation}
    \label{eq:constellation}
    \rho(\anAFsymbol') = \bigotimes_{\arga \in \setargs'} \rho(\arga) ~\otimes~ \bigotimes_{\arga \in \setargs\setminus\setargs'} \overline{\rho(\arga)}
\end{equation}

The constellation probability of an argument $\arga$ is then defined as follows.

\begin{definition}
  \label{def:probqueryconstellation}
  Given a probabilistic graph $\tuple{\setargs, \setattacks,
    \rho}$ over an argumentation framework $\anAFsymbol =
  \tuple{\setargs, \setattacks}$,  an argument $\arga \in \setargs$, a semantics $\gensem
\in \set{\CF, \AD, \CO, \GR, \ST, \PR}$, and a probability distribution over subgraphs $\rho: \wp(\anAFsymbol) \to [0,1]$:
\begin{equation}
  \label{eq:probargc}
  \mbox{\textbf{PROB-C}}(\arga, \gensem, \anAFsymbol) = \bigoplus_{\set{\anAFsymbol' \in \wp(\anAFsymbol) ~\mid~ E \in \setgenext{\gensem}{\anAFsymbol'} \mbox{ and } \arga \in E}} \rho(\anAFsymbol').
\end{equation}
\end{definition}

In particular, let us consider the subclass $\mbox{\textbf{PROB-C}}^{\mbox{IND}}(\arga, \gensem, \anAFsymbol)$ where the $\rho: \wp(\anAFsymbol) \to [0,1]$ is \eqref{eq:constellation}. For it, we can prove that $\mbox{\textbf{PROB-C}}^{\mbox{IND}}(\arga, \gensem,
\anAFsymbol)$ is an instance of AMC.\footnote{This results has been---implicitly---already provided in \cite{bistarelli_ProbabilisticArgumentationFrameworks_18} as the authors used ProbLog to solve $\mbox{\textbf{PROB-C}}^{\mbox{IND}}(\arga, \gensem,
\anAFsymbol)$ and, from \cite{Fierens2015,kimmig_Algebraicmodelcounting_17}, it is known that AMC generalises ProbLog inferences.}
  
\begin{theorem}
  \label{thm:amcconst}
  Given a probabilistic graph $\tuple{\setargs, \setattacks, \rho}$
  over an argumentation framework $\anAFsymbol = \tuple{\setargs,
    \setattacks}$, and given $\arga \in \setargs$, 
  AMC generalises $\mbox{\textbf{PROB-C}}^{\mbox{IND}}(\arga, \gensem,
  \anAFsymbol)$.

  \begin{proof} (Sketch.)
    From \eqref{eq:constellation}, given $\anAFsymbol'  = \tuple{\setargs', \setattacks'} \sqsubseteq \anAFsymbol = \tuple{\setargs, \setattacks}$, we need propositional theories with models of the form
    $\Phi(\anAFsymbol') = \bigwedge_{\arga \in \setargs'} \arga \land \bigwedge_{\arga \in \setargs\setminus\setargs'} \lnot \arga.$
    \eqref{eq:probargc} then requires to consider models $\Phi(\anAFsymbol')$ for which the query argument belong to an extension, \ie\
        $\Phi^C(\anAFsymbol, \gensem, \arga) = \linebreak \bigvee_{\set{\anAFsymbol' \in \wp(\anAFsymbol) ~\mid~ E \in \setgenext{\gensem}{\anAFsymbol'} \mbox{ and } \arga \in E}} \Phi(\anAFsymbol').$\qedhere
  \end{proof}
\end{theorem}

Taking into account Theorems \ref{thm:amcarg} and \ref{thm:amcconst}, it follows that---in general---$\mbox{\textbf{PROB}}(\arga, \gensem,
  \anAFsymbol)$ is not equivalent to $\mbox{\textbf{PROB-C}}^{\mbox{IND}}(\arga, \gensem,
  \anAFsymbol)$.
  
\begin{proposition}
There are probabilistic graphs $\tuple{\setargs, \setattacks, \rho}$
  over an argumentation framework $\anAFsymbol = \tuple{\setargs,
    \setattacks}$, for which given $\arga \in \setargs$ and a semantics $\gensem$, 
  $\mbox{\textbf{PROB}}(\arga, \gensem,
  \anAFsymbol) \neq \mbox{\textbf{PROB-C}}^{\mbox{IND}}(\arga, \gensem,
  \anAFsymbol)$.
\begin{proof}
  Let us consider $\anAFsymbol = \tuple{\set{\arga, \argb, \argc}, \set{\tuple{\arga, \argb}, \tuple{\argb, \argc}}}$, $\rho(\arga) = \weight{\arga} \in [0,1]$, $\rho(\argb) = \weight{\argb} \in [0,1]$, $\rho(\argc) = \weight{\argc} \in [0,1]$, and $\overline{\weight{\cdot}} = (1 - \weight{\cdot})$.
$\mbox{\textbf{PROB}}(\argc, \GR,
  \anAFsymbol)  =   \weight{\arga}  \cdot \overline{\weight{\argb}}  \cdot  \weight{\argc}$ while
  $\mbox{\textbf{PROB-C}}^{\mbox{IND}}(\argc, \GR,
  \anAFsymbol) = (\weight{\arga}  \cdot \weight{\argb}  \cdot \weight{\argc}) + (\weight{\arga}  \cdot \overline{\weight{\argb}}  \cdot \weight{\argc}) + (\overline{\weight{\arga}}  \cdot \overline{\weight{\argb}}  \cdot \weight{\argc})$.
\end{proof}
\end{proposition}

\subsection{Argumentative Reasoning over Beta Distributions}
\label{sec:distributionsemanticsbeta}

\begin{table*}[]
    \centering
    \caption{Probabilistic inferences over $\anAFsymbol_E$, with chosen $\rho(q)$ labels, using Definitions \ref{def:probquery} and \ref{def:probqueryconstellation}.}
    \begin{tabular}{l p{3.5cm} p{3.5cm} p{3.5cm}}
    \toprule
    $q$ & $\rho(q)$ & $\mbox{\textbf{PROB}}(q, \AD,
  \anAFsymbol_E)$ & $\mbox{\textbf{PROB-C}}^{\mbox{IND}}(q, \AD,
  \anAFsymbol_E)$\\
    \midrule
    \arga 
    & $\dbeta(1.00, 1.00)$ \newline
    \textit{Chances about even} with \textit{no confidence}
    & $\dbeta(1.35, 2.07)$ \newline
    \textit{Somewhat unlikely} with \textit{low confidence}
    & $\dbeta(1.14, 1.35)$ \newline
    \textit{Chances about even} with \textit{some confidence}
    \\
    \argb 
    & $\dbeta(17.00, 2.00)$ \newline
    \textit{Very likely} with \textit{high confidence}
    & $\dbeta(14.57, 6.06)$ \newline
    \textit{Likely} with \textit{high confidence}
    & $\dbeta(2.41, 1.97)$ \newline
    \textit{Somewhat likely} with \textit{some confidence}
    \\
    \argc 
    & $\dbeta(4.00, 15.00)$ \newline
    \textit{Unlikely} with \textit{high confidence}
    & $\dbeta(1.00, +\infty)$ \newline
    \textit{Absolutely not likely} with \textit{total confidence}
    & $\dbeta(1.03, 92.30)$\newline
    \textit{Very unlikely} with \textit{total confidence}
    \\
    \argd 
    & $\dbeta(5.00, 1.50)$ \newline
    \textit{Likely} with \textit{some confidence}
    & $\dbeta(7.05, 5.21)$ \newline
    \textit{Somewhat likely} with \textit{some confidence}
    & $\dbeta(5.20, 1.64)$ \newline
    \textit{Likely} with \textit{some confidence}
    \\
    \bottomrule\\
    \end{tabular}
    \label{tab:results}
\end{table*}

Let us consider again our running example (Example \ref{ex:ex}) and let us suppose to label arguments with the beta distributions\footnote{In \cite{santini_AreMyArguments_18}, the authors also label arguments with beta distributions in the form of subjective logic opinions \cite{josang_Subjectivelogic_16} for representing their level of trust.} as listed in the second column of Table \ref{tab:results}.
Table \ref{tab:results} also shows the results of probabilistic inferences over $\anAFsymbol_E$ when considering both $\mbox{\textbf{PROB}}$ (Definition \ref{def:probquery}) and $\mbox{\textbf{PROB-C}}^{\mbox{IND}}$ (Definition \ref{def:probqueryconstellation}) using $\AD$ in both cases as argumentation semantics, and the machinery provided by \cite{cerutti_HandlingEpistemicAleatory_22}, recalled in Section \ref{sec:mach22}.

When using $\mbox{\textbf{PROB}}$ (Definition \ref{def:probquery}), the knowledge we inject through the structure of the argumentation graph heavily affects the result of probabilistic query. Because there is no admissible extension in which $\argc$ is accepted, its probabilistic assessment is \textit{absolutely not likely} with \textit{total confidence} (equivalent to $e^{\oplus}$). Moreover, the probabilistic assessment of $\arga$ and $\argb$ considers the fact that admissible sets are not complete, hence there are cases where only one of the two (alternating) are in such a set. Finally, we begun our example from $\argd$ and its uncertainty assessment as \emph{likely} with \emph{some confidence}, and post-hoc we assumed both $\anAFsymbol_E$ as an argumentation graph representing our understanding of the relationships between possible relevant arguments, and the labels listed in the second column of Table \ref{tab:results} which complete the label for $\argd$ we already had. Such assumptions affects $\argd$ which now becomes \emph{somewhat likely} with \emph{some confidence}.
The aleatory evaluation of $\argd$ has been affected by the uncanny uninformative $\rho(\arga)$, despite being unattacked in $\anAFsymbol_E$. In future work, we will characterise possible \textit{rationality constraints} building upon the extensive literature on epistemic approach to probabilistic argumentation \cite{thimm_probabilisticsemanticsabstract_12} and ranking semantics, e.g., \cite{ranking}.
$\mbox{\textbf{PROB}}$ can thus become a useful test for assessing the consistency of both probabilistic assessments and the logical dependencies between arguments.

When using $\mbox{\textbf{PROB-C}}^{\mbox{IND}}$ (Definition \ref{def:probqueryconstellation}), instead, the assessment considers whether a rational agent is aware of the presence of arguments. 
The most striking difference with $\mbox{\textbf{PROB}}$ concerns $\argb$ which, from \textit{very likely} with \textit{high confidence} then becomes \textit{somewhat likely} with \textit{some confidence}.
This is once again due to the fact that $\arga$, despite being unattacked, has such an uncanny uninformative assessment which affects the overall computation irrespectively whether \arga\ is considered in the subgraphs or not. Indeed, $\overline{\dbeta(1.00, 1.00)} = \dbeta(1.00, 1.00)$.

\section{Conclusions}
\label{sec:conclusions}

In this paper, we show a machinery for effective evaluation of argumentation frameworks with both epistemic and aleatory uncertainty, represented through beta distributions, which are used in intelligence analysis and scientific assessments. In particular, we focused on the machinery provided by (algebraic) model counting, which led us to introduce a novel probabilistic evaluation ($\mbox{\textbf{PROB}}$, Definition \ref{def:probquery}) which differ from existing probabilistic evaluations based upon the constellation approach ($\mbox{\textbf{PROB-C}}^{\mbox{IND}}$ Definition \ref{def:probqueryconstellation}). We nevertheless show how both probabilistic evaluations are instances of the model counting problem, and illustrated their difference in our running example.

The paper provides some basic preliminary results which may enable further developments and raises more questions than provides solutions. In particular, while we prove the connection between probabilistic logical inferences and probabilistic argumentation, the results of our running example (Table \ref{tab:results}) invite discussions on a wide range of topics.

First of all, there is an underlying independence assumption between the probabilistic labels associated to each argument, e.g., $\rho(\arga \mid \argb) = \rho(\arga)$ for any $\arga, \argb$. From one hand this is unrealistic: there is always the possibility of having confounding variables. At the same time, experimental analysis on synthetic data \cite{kaplan_TrustEstimationSources_18} show that 
dependencies often do not change significantly a probabilistic assessment. Deciding whether the independence assumption between random variables is reasonable or not appears thus to be debatable, and it should be represented in an argumentative assessment of a phenomenon, thus adding a meta-level.

Definition \ref{def:probquery} can also be seen as a way to \textit{weight} argumentation extensions. From this perspective, it goes in the direction presented, for instance, in \cite[\S\ 3.2]{toniolo_EnumeratingPreferredExtensions_17} and \cite[\S\ 3]{thimm_ProbabilitiesExtensionsAbstract_17}. While our focus is on supporting probabilistic reasoning, we will analyse the connection further in future work.


In this paper, we limited ourselves to consider labels for arguments only. We, however, can think of extending the approach considering uncertain probabilities labelling for attacks too thanks to formal results provided in \cite{Baroni2011b}. In that paper, the authors introduced the argumentation framework with recursive attacks and, among other results, \cite[Def. 19]{Baroni2011b} shows how to represent attacks as arguments while ensuring \cite[Prop. 6--12]{Baroni2011b} semantic correspondence. 

Finally, given the connection between probabilistic inferences and Bayesian networks \cite{darwiche_differentialapproach_03}, an interesting future research question concerns the syntactic relationship---if possible---between Bayesian networks with binary random variables and argumentation frameworks which ensures semantic correspondence too. 

\section*{Acknowledgements}
We thank Matthias Thimm for his valuable comments on an early version of this manuscript.

\bibliographystyle{vancouver}
\bibliography{biblio,additional}

\begin{thebibliography}{10}

\bibitem{ipcc_ClimateChange2022_NaN}
{IPCC}.
\newblock Climate {{Change}} 2022: {{Impacts}}, {{Adaptation}}, and
  {{Vulnerability}}.
\newblock {Cambridge University Press}; In press.

\bibitem{HORA1996217}
Hora SC.
\newblock Aleatory and Epistemic Uncertainty in Probability Elicitation with an
  Example from Hazardous Waste Management.
\newblock Reliability Engineering \& System Safety. 1996;54(2):217-23.

\bibitem{Laplace-prob}
Laplace PS.
\newblock A {{Philosophical Essay}} on {{Probabilities}}.
\newblock {Springer}; 1825.

\bibitem{cerutti_EvidentialReasoningLearning_22}
Cerutti F, Kaplan L, Sensoy M.
\newblock Evidential {{Reasoning}} and {{Learning}}: A {{Survey}}.
\newblock In: {{IJCAI}} 2022; 2022. .

\bibitem{josang_Subjectivelogic_16}
J{\o}sang A.
\newblock Subjective Logic: A Formalism for Reasoning under Uncertainty.
\newblock {Springer}; 2016.

\bibitem{sato:iclp95}
Sato T.
\newblock A Statistical Learning Method for Logic Programs with Distribution
  Semantics.
\newblock In: Proceedings of the 12th {{International Conference}} on {{Logic
  Programming}} ({{ICLP-95}}); 1995. .

\bibitem{Fierens2015}
Fierens D, {den Broeck} G, Renkens J, Shterionov D, Gutmann B, Thon I, et~al.
\newblock Inference and Learning in Probabilistic Logic Programs Using Weighted
  {{Boolean}} Formulas.
\newblock Theory and Practice of Logic Programming. 2015 May;15(03):358-401.

\bibitem{kimmig_Algebraicmodelcounting_17}
Kimmig A, {Van den Broeck} G, De~Raedt L.
\newblock Algebraic Model Counting.
\newblock Journal of Applied Logic. 2017 Jul;22:46-62.

\bibitem{Kaplan2018}
Kaplan L, Ivanovska M.
\newblock Efficient Belief Propagation in Second-Order {{Bayesian}} Networks
  for Singly-Connected Graphs.
\newblock International Journal of Approximate Reasoning. 2018 Feb;93:132-52.

\bibitem{cerutti_ProbabilisticLogicProgramming_19}
Cerutti F, Kaplan L, Kimmig A, Sensoy M.
\newblock Probabilistic {{Logic Programming}} with {{Beta-Distributed Random
  Variables}}.
\newblock In: {{AAAI}}; 2019. p. 7769-76.

\bibitem{cerutti_HandlingEpistemicAleatory_22}
Cerutti F, Kaplan LM, Kimmig A, Sensoy M.
\newblock Handling {{Epistemic}} and {{Aleatory Uncertainties}} in
  {{Probabilistic Circuits}}.
\newblock Machine Learning. 2022.

\bibitem{li_Relaxingindependenceassumptions_11}
Li H, Oren N, Norman TJ.
\newblock Relaxing Independence Assumptions in Probabilistic Argumentation.
\newblock In: {{ArgMAS}}; 2011. .

\bibitem{hunter_probabilisticapproachmodelling_13}
Hunter A.
\newblock A Probabilistic Approach to Modelling Uncertain Logical Arguments.
\newblock International Journal of Approximate Reasoning. 2013 Jan;54(1):47-81.

\bibitem{hunter_Probabilisticqualificationattack_14}
Hunter A.
\newblock Probabilistic Qualification of Attack in Abstract Argumentation.
\newblock International Journal of Approximate Reasoning. 2014
  Jan;55(2):607-38.

\bibitem{mastrandrea_IPCCAR5guidance_11}
Mastrandrea MD, Mach KJ, Plattner GK, Edenhofer O, Stocker TF, Field CB, et~al.
\newblock The {{IPCC AR5}} Guidance Note on Consistent Treatment of
  Uncertainties: A Common Approach across the Working Groups.
\newblock Climatic Change. 2011 Aug;108(4):675.

\bibitem{murphy_Machinelearningprobabilistic_12}
Murphy KP.
\newblock Machine Learning: A Probabilistic Perspective.
\newblock {MIT press}; 2012.

\bibitem{Darwiche2004}
Darwiche A, {Darwiche}, {Adnan}.
\newblock New {{Advances}} in {{Compiling CNF}} to {{Decomposable Negation
  Normal Form}}.
\newblock In: Proceedings of the 16th {{European Conference}} on {{Artificial
  Intelligence}}. {NLD}: {IOS Press}; 2004. p. 318-22.

\bibitem{darwiche_KnowledgeCompilationMap_02}
Darwiche A, Marquis P.
\newblock A {{Knowledge Compilation Map}}.
\newblock J Artif Int Res. 2002 Sep;17(1):229-64.

\bibitem{darwiche_differentialapproach_03}
Darwiche A.
\newblock A Differential Approach to Inference in {{Bayesian}} Networks.
\newblock Journal of the ACM. 2003 May;50(3):280-305.

\bibitem{dung_acceptabilityarguments_95}
Dung PM.
\newblock On the Acceptability of Arguments and Its Fundamental Role in
  Nonmonotonic Reasoning, Logic Programming and n-Person Games.
\newblock Artificial Intelligence. 1995 Sep;77(2):321-57.

\bibitem{besnard_Checkingacceptabilityset_04}
Besnard P, Doutre S.
\newblock Checking the Acceptability of a Set of Arguments.
\newblock In: {{NMR04}}; 2004. .

\bibitem{Darwiche2011}
Darwiche A.
\newblock {{SDD}}: {{A New Canonical Representation}} of {{Propositional
  Knowledge Bases}}.
\newblock In: Proceedings of the {{Twenty-Second International Joint
  Conference}} on {{Artificial Intelligence}} - {{Volume Volume Two}}. {AAAI
  Press}; 2011. p. 819-26.

\bibitem{totis_SMProbLogStableModel_21a}
Totis P, Kimmig A, De~Raedt L.
\newblock {{SMProbLog}}: {{Stable Model Semantics}} in {{ProbLog}} and Its
  {{Applications}} in {{Argumentation}}.
\newblock In: {{IJCLR}} 2021; 2021. .

\bibitem{faber_SolvingSetOptimization_16}
Faber W, Vallati M, Cerutti F, Giacomin M.
\newblock Solving {{Set Optimization Problems}} by {{Cardinality Optimization}}
  with an {{Application}} to {{Argumentation}}.
\newblock In: {{ECAI}} 2016 - 22nd {{European Conference}} on {{Artificial
  Intelligence}}, 29 {{August-2 September}} 2016, {{The Hague}}, {{The
  Netherlands}} - {{Including Prestigious Applications}} of {{Artificial
  Intelligence}} ({{PAIS}} 2016); 2016. p. 966-73.

\bibitem{pipatsrisawat_SolvingWeightedMaxSAT_08}
Pipatsrisawat K, Palyan A, Chavira M, Choi A, Darwiche A.
\newblock Solving {{Weighted Max-SAT Problems}} in a {{Reduced Search Space}}:
  {{A Performance Analysis}}.
\newblock Journal on Satisfiability, Boolean Modeling and Computation. 2008
  Jan;4(2-4):191-217.

\bibitem{bistarelli_ProbabilisticArgumentationFrameworks_18}
Bistarelli S, Mantadelis T, Santini F, Taticchi C.
\newblock Probabilistic {{Argumentation Frameworks}} with {{MetaProbLog}} and
  {{ConArg}}.
\newblock In: 2018 {{IEEE}} 30th {{International Conference}} on {{Tools}} with
  {{Artificial Intelligence}} ({{ICTAI}}); 2018. p. 675-9.

\bibitem{santini_AreMyArguments_18}
Santini F, J{\o}sang A, Pini MS.
\newblock Are {{My Arguments Trustworthy}}? {{Abstract Argumentation}} with
  {{Subjective Logic}}.
\newblock In: 2018 21st {{International Conference}} on {{Information Fusion}}
  ({{FUSION}}); 2018. p. 1982-9.

\bibitem{thimm_probabilisticsemanticsabstract_12}
Thimm M.
\newblock A Probabilistic Semantics for Abstract Argumentation.
\newblock In: Proceedings of the 20th European Conference on Artificial
  Intelligence. {{ECAI}}'12. {NLD}: {IOS Press}; 2012. p. 750-5.

\bibitem{ranking}
Bonzon E, Delobelle J, Konieczny S, Maudet N.
\newblock A Comparative Study of Ranking-Based Semantics for Abstract
  Argumentation.
\newblock In: Schuurmans D, Wellman MP, editors. Proceedings of the Thirtieth
  {AAAI} Conference on Artificial Intelligence, February 12-17, 2016, Phoenix,
  Arizona, {USA}. {AAAI} Press; 2016. p. 914-20.
\newblock Available from:
  \url{http://www.aaai.org/ocs/index.php/AAAI/AAAI16/paper/view/12465}.

\bibitem{kaplan_TrustEstimationSources_18}
Kaplan L, {\c S}ensoy M.
\newblock Trust {{Estimation}} of {{Sources Over Correlated Propositions}}.
\newblock In: 2018 21st {{International Conference}} on {{Information Fusion}}
  ({{FUSION}}); 2018. p. 414-21.

\bibitem{toniolo_EnumeratingPreferredExtensions_17}
Toniolo A, Norman TJ, Oren N.
\newblock Enumerating {{Preferred Extensions}}: {{A Case Study}} of {{Human
  Reasoning}}.
\newblock In: Theory and {{Applications}} of {{Formal Argumentation}} - 4th
  {{International Workshop}}, {{TAFA}} 2017, {{Melbourne}}, {{VIC}},
  {{Australia}}, {{August}} 19-20, 2017, {{Revised Selected Papers}}; 2017. p.
  192-210.

\bibitem{thimm_ProbabilitiesExtensionsAbstract_17}
Thimm M, Baroni P, Giacomin M, Vicig P.
\newblock Probabilities on {{Extensions}} in {{Abstract Argumentation}}.
\newblock In: Theory and {{Applications}} of {{Formal Argumentation}} - 4th
  {{International Workshop}}, {{TAFA}} 2017, {{Melbourne}}, {{VIC}},
  {{Australia}}, {{August}} 19-20, 2017, {{Revised Selected Papers}}; 2017. p.
  102-19.

\bibitem{Baroni2011b}
Baroni P, Cerutti F, Giacomin M, Guida G.
\newblock {{AFRA}}: {{Argumentation}} Framework with Recursive Attacks.
\newblock International Journal of Approximate Reasoning. 2011;52(1):19-37.

\end{thebibliography}

\end{document}